\documentclass{article}

\usepackage[super,numbers]{natbib}
\usepackage[accepted]{nyas}

\usepackage{graphicx}

\begin{document}

\title{Can Large Language Models Reason and Plan?}

\author{Subbarao Kambhampati\\
School of Computing \& Augmented Intelligence\\
Arizona State University\\
email: rao@asu.edu}

\date{A version appears in the Annals of The New York Academy of Sciences: https://nyaspubs.onlinelibrary.wiley.com/doi/10.1111/nyas.15125}

\maketitle



\begin{figure}[t]
    \centering
    \includegraphics[width=\linewidth]{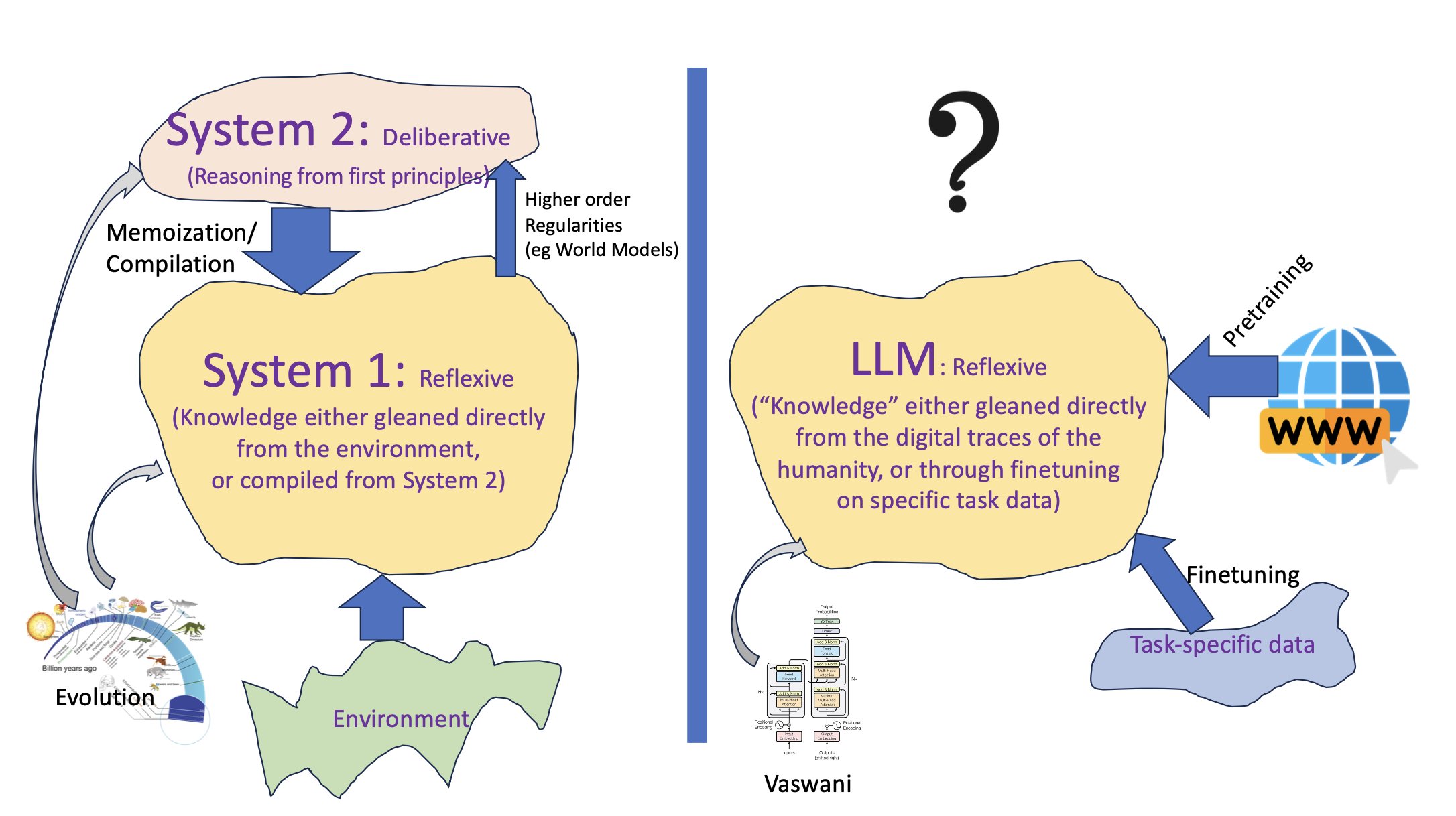}
    \caption{An informal account of viewing LLM as a giant external non-veridical memory that acts as a pseudo System 1}
    \label{fig:sys12}
\end{figure}

Large Language Models (LLMs), essentially n-gram models on steroids\footnote{LLMs are trained to predict the distribution of the n-th token given n-1 previous tokens. GPT3.5 that powered the original ChatGPT is, for example, a roughly 3001-gram model of language.} that have been trained on web-scale language corpora
(or, effectively, our civilizational knowledge), 
have caught our collective imagination
with linguistic behaviors that no one expected text completion systems to possess \cite{rao-cacm-llm}.
By training and operation, LLMs are perhaps best seen as giant non-veridical memories akin to an external System 1 \cite{thinking-fast-slow}  for us all (see Figure~\ref{fig:sys12}).
Their seeming versatility has however led many researchers to wonder whether they can also do well on planning and reasoning tasks typically associated with System 2 competency. 






 Nothing in the training and use of LLMs would seem to suggest remotely  that they can do any type of principled reasoning (which, as we know, often involves computationally hard inference/search). 
 What LLMs are good at is a form of universal approximate retrieval.
 Unlike databases that index and retrieve data exactly, LLMs, as n-gram models, probabilistically reconstruct completions for the prompt word by word--a process we shall refer to as {\em approximate retrieval}.
 This means that LLMs can't even guarantee memorizing complete answers, something that is the flip side of their appeal about constructing ``novel'' prompt completions on the fly.
 The boon (``creativity") and bane (``hallucination") of LLMs is that n-gram models will naturally mix and match--and have almost as much trouble strictly memorizing as we do.  It is indeed the very basis of their appeal.
 %

 %
 Despite this, the ``Large Language Models are Zero-Shot $\langle$insert-your-reasoning-task $\rangle$'' has almost become a meme paper title! 
 At some level, this trend is perhaps inevitable as in the era of LLMs, AI has become a form of ersatz natural science \cite{rao-ersatz}–driven by observational studies of capabilities of these behemoth systems. 


So, are these n-gram models on steroids really capable of planning and reasoning? In the summer of 2022, when my research group wanted to better answer this question, most reasoning claims were still somewhat anecdotal. So, we set out to evaluate GPT3 on a set of planning instances derived from the domains typically used in the International Planning Competition (IPC) –including the well known Blocks World\footnote{{\tt https://en.wikipedia.org/wiki/Blocks\_world}}. Our results \cite{valmeekam-still-cant} were contrary to the anecdotal claims about the planning abilities of LLMs, and when we made them public,  received significant attention in the AI circles. 

By the beginning of 2023, with the wide-spread public release of ChatGPT, and later, GPT4, there were a slew of additional claims, including in refereed papers, about LLM’s abilities to reason and plan. So we decided to repeat our tests on both GPT3.5 and GPT4 \cite{valmeekam2023on}. Initial results showed that there was some improvement in the accuracy of generated plans from GPT3 to GPT3.5 to GPT4, with GPT4 reaching 30\% empirical accuracy in the Blocks World (albeit still lower in other domains). We then wanted to know whether the modest improvement  is because of the improved approximate retrieval abilities or whether GPT4 is actually doing/searching for plans. 

Let us pause to note that my interest here is not whether LLMs can fake reasoning (by giving correct answers to reasoning tasks from memory and pattern finding), but whether they can actually do principled reasoning. Of course, seeing patterns in reasoning problems is not anything to be sneezed at. After all, our interest in mastering it is what is behind much of ``street fighting" math (e.g. George P\'olya's "How to Solve it").  But finding approximate shortcuts over provably correct reasoning procedures is obviously not equivalent to doing reasoning--unless you have an ability to establish from first principles  that your hunch is actually correct. It  is challenging to decide whether a system (or a human, for that matter) is memorizing or solving a problem from scratch–especially as the systems (or humans) get trained on larger and larger  “question banks.” This is a challenge that most instructors and interviewers are acutely aware of. Think of that infamous “Why are manhole covers round?” interview question. While it may well have given the interviewer an insight into the candidate’s analytical reasoning skills the very first time it was asked, all it does with high probability now is to confirm whether the candidate trained on the interview question banks!

Considering that the LLMs don’t suffer some of the normal limitations of humans–such as having a life on the side, and thus not having the time or  inclination to focus exclusively on the test/interview preparation for long periods, they can support approximate retrieval over webscale corpora.  My research group wanted to check if the improved performance of GPT4 is because of approximate retrieval from a larger training corpus, or really comes from its ability to plan. One way of checking this for planning tasks is to reduce the effectiveness of approximate retrieval by obfuscating the names of the actions and objects in the planning problem. When we did this for test domains \cite{valmeekam2023on,valmeekam2023planbench}, GPT4’s empirical performance plummeted precipitously, despite the fact that none of the standard off-the-shelf AI planners have any trouble with such obfuscation.\footnote{As these results came about at the height of  sparks of AGI/existential risk angst, we couldn’t resist the tongue-in-cheek editorializing that  if GPT4 ever goes rogue, you can stymie it by throwing  a simple planning problem at it! Humor aside, nothing in our studies showed that GPT4 is capable of generating executable plans autonomously.}

Perhaps they can’t do planning autonomously straight out of the box, but can they do it with a little nudge? There are broadly two popular techniques for such nudging.  The first, called “fine tuning,” is rather straightforward: take a general LLM and fine tune it on planning problems (i.e., instances and their solutions), with the hope that they will subsequently make better guesses (see the left-hand side of Figure~\ref{fig:sys12}).
While our own limited experiments didn’t show any significant improvement through fine tuning, it is possible that with even more fine tuning data and effort, the quality of LLM guesses may well improve. But all that such fine tuning is doing is converting the planning task into a memory-based approximate retrieval (akin to the memorization/compilation from System 2 to System 1; see Figure~\ref{fig:sys12}). It doesn’t prove that LLMs are able to plan.

The second way to improve planning (and reasoning) performance is to prompt an LLM back with hints/suggestions about how it can improve its initial plan guess. The crucial questions here are (a)  whether this back prompting is manual or automated (b) who is certifying the correctness of the final answer and (c) whether the prompts inject additional problem knowledge or are just merely exhorting the LLM to try again.

The cleanest approach–one we advocate\cite{valmeekam2023on,rao-llm-modulo}–is to let an external model-based plan verifier do the back prompting and to certify the correctness of the final solution. In general, such {\em LLM-Modulo} frameworks\cite{rao-llm-modulo} can gainfully leverage the amazing idea generation capabilities of LLMs with sound external verifiers in a generate-test-critique framework with gurantees. 

\begin{figure}
    \centering
    \includegraphics[width=\linewidth]{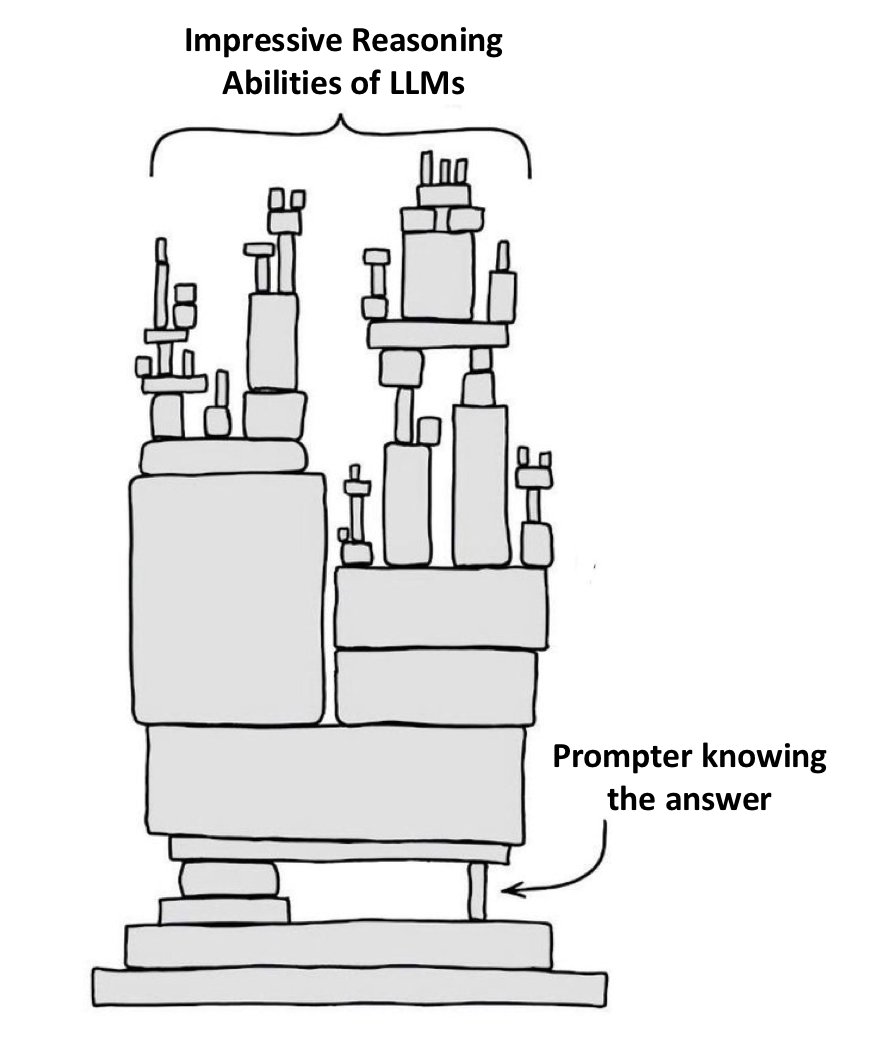}
    \caption{Claimed reasoning capabilities of LLMs are sometimes due to the subconscious helpful iterative prompting by the humans in the loop (graphic adapted from https://xkcd.com/2347/ under Creative Commons License)}
    \label{fig:llm-reasoning}
\end{figure}

In contrast, by far the more popular methodology is to have the human in the loop prompt the LLM iteratively. 
The problem with this is that it is highly susceptible to the Clever Hans effect\footnote{https://en.wikipedia.org/wiki/Clever\_Hans}, where the LLM is merely generating guesses, and it is the human in the loop, with the knowledge of right vs. wrong solutions, who is steering the LLM–even if they didn’t set out to do so deliberately. The credit and blame for the ensuring accuracy, if any, falls squarely on the human in the loop. The relevance of such a framework becomes questionable when the human-in-the-loop doesn’t know (or is unable to verify) the answer to the reasoning/planning problem themselves. Thus the tongue-in-cheek characterization of LLM reasoning abilities in Figure~\ref{fig:llm-reasoning}.

A variation on the second approach is to have the LLM itself “critique” the guesses it generates and iteratively self-improve. Although some papers seem to swear by such a  “self-improvement” capability of LLMs,  the plausibility of such a claim  hinges on the belief that the LLMs are better at verifying their solutions than they are at generating them. While never explicitly justified, the assumption rests on either analogies to humans or indirect nods to computational complexity arguments. While humans sometimes do show the capability of correcting their own erroneous guesses with self-critiquing, there seems to be no basis for that assumption in the case of LLMs. And while for many computational tasks (e.g. those in class NP\footnote{NP stands for {\em nondeterministic polynomial}, and covers the class of computational problems whose solutions can be verified in polynomial time.}), the verification is often of lower complexity than generation, that fact doesn’t seem particularly  relevant for LLMs which are generating (approximately retrieving) guesses, rather than actually solving the problem with guarantees. 
%
%
Indeed two recent studies from my lab--one on plan verification \cite{valmeekam2023can} and the other on constraint verification \cite{stechly2023gpt}--seem to throw cold water on this optimism by showing that with ``self-verification" performance actually worsens. This is because LLMs hallucinate both false positives and false negatives while verifying the solutions they generate. 
One reason this is not recognized in earlier literature is that there 
self-verification claims are often made in the context of tacit knowledge tasks for which there is little possibility of a verifier (e.g. writing/improving essays), making it harder to evaluate whether LLM's critiquing actually helped. Paradoxically, the fact that it is infeasible to write sound verifiers for tacit knowledge tasks also makes it easier to mistake LLMs for being as reasonable a critic as any!\footnote{In other words, LLMs can be as good as that Peloton instructor in confidently critiquing Christopher Nolan movies.}
In other cases, an external simulator winds up playing the role of sound verification. 

While the foregoing questions the claims that LLMs are capable of planning/reasoning, it is not meant to imply  that LLMs don’t have any constructive roles to play in solving planning/reasoning tasks. In particular, their uncanny ability to generate ideas/potential candidate solutions–albeit with no guarantees about those guesses–can still be valuable in the  “LLM-Modulo”  setups\cite{rao-llm-modulo},  in conjunction with either model-based verifiers or expert humans in the loop. 
The trick to avoiding ascribing autonomous reasoning capabilities to LLMs is to recognize that LLMs are generating potential answers that still need to be checked by
external verifiers. 


The skeptical reader might now ask: But what about all those papers at high profile AI conferences that claim to show planning abilities of LLMs? To analyze those claims, we need to first  understand that solving planning tasks requires (a) having the necessary planning domain knowledge–the actions and their preconditions, effects; the standard hierarchical recipes (e.g. task reduction schemas in Hierarchical Task Network planning),
past cases/plans etc. and (b) being able to assemble this knowledge into an executable plan that takes care of any subgoal/resource interactions. The first can be called the knowledge acquisition part, and the second reasoning/planning part.  Many of the papers claiming planning abilities of LLMs, on closer examination, wind up confusing general planning knowledge extracted from the LLMs for executable plans. When all we are looking for are abstract plans, such as “wedding plans,” with no intention of actually executing said plans directly, it is easy to confuse them for complete executable plans. Indeed, our close examination of several papers claiming planning capabilities \cite{llm-tutorial} for LLMs suggests that they either are evaluating in domains/tasks where subgoal interactions can be safely ignored, or  delegating the interaction resolution (reasoning)  to the humans in the loop (who, through repeated prompting, have to “correct” the plan). Sometimes, in common sense domains, or with enough fine tuning, the “assembling” part may also be obviated by having seen a case that pretty much corresponds to the problem that needs to be solved. Without these assumptions or mitigations, the plans that come out of LLMs may look reasonable to the lay user, but lead to execution time interactions and errors. These issues are illustrated in part  by a recent news story about the proliferation of travel planning books \cite{nyt-travel-books}, mostly auto-extracted from LLMs, and the ensuing disappointment of the unsuspecting end users who buy them mistaking them for usable plans!



\begin{figure}[t]
    \centering
    \includegraphics[width=\linewidth]{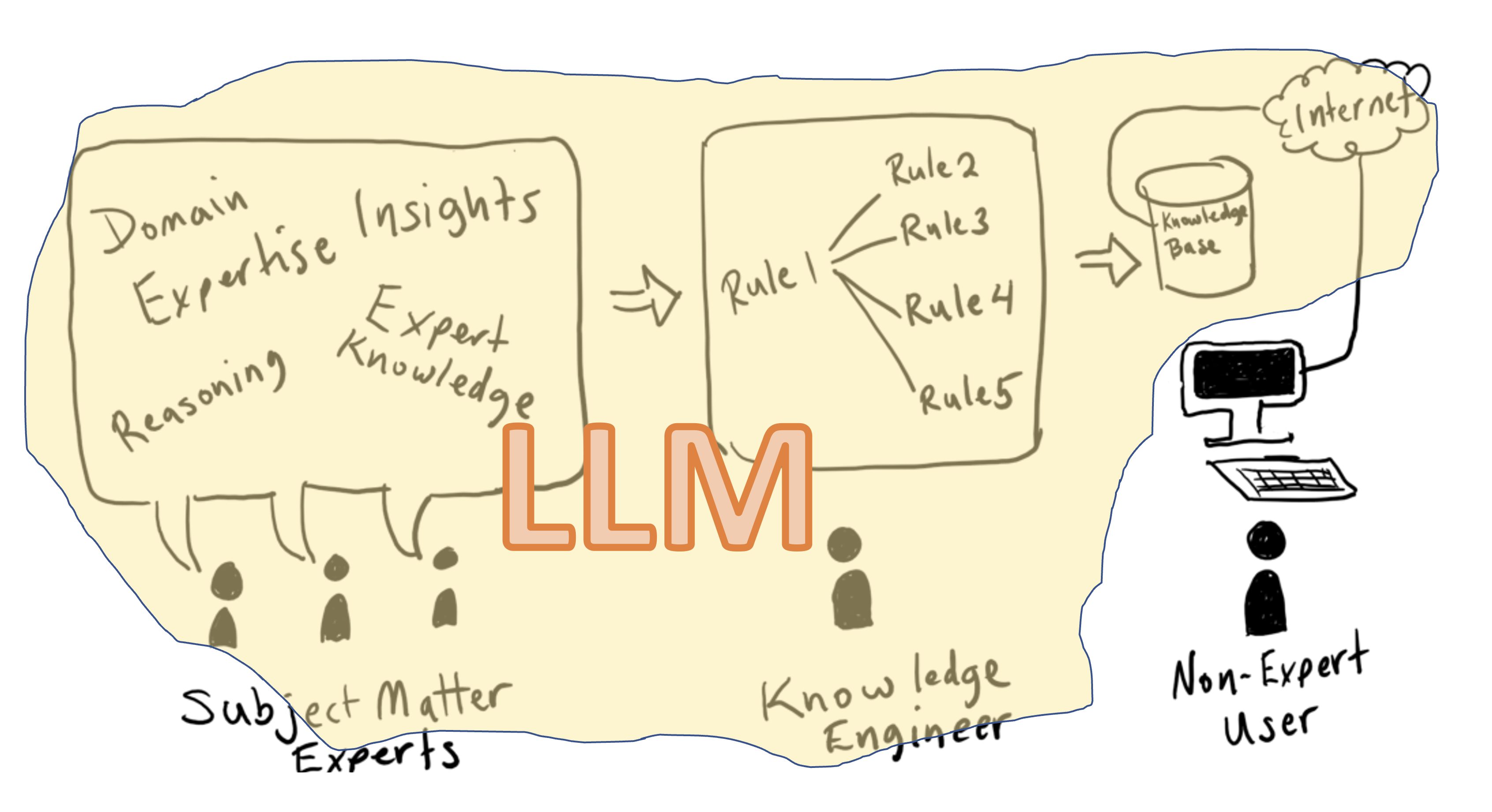}
    \caption{Viewing LLMs as an approximate knowledge source trained over civilizational knowledge}
    \label{fig:avenging-polanyi}
\end{figure}

The fact that LLMs are often good at extracting planning knowledge can indeed be gainfully leveraged. As we have  argued in recent work \cite{guan2023leveraging}, LLMs can  be a rich source of approximate models of world/domain dynamics and user preferences, as long as the humans (and any specialized critics) in the loop verify and refine the models, and give them over to model-based solvers. This way of using LLMs has the advantage that the humans need only be present when the dynamics/preference model is being teased out and refined, with the actual planning after that being left to sound frameworks with correctness guarantees\cite{rao-llm-modulo}. 

Such a framework has striking similarities to knowledge-based AI systems of yore, with  LLMs  effectively replacing the ``knowledge engineer'' (Figure~\ref{fig:avenging-polanyi}). Given the rather quixotic and dogmatic shift of AI away from approaches that accept domain knowledge from human experts, something I bemoaned in ``{\em P\'olanyi’s revenge and AI's new
romance with tacit knowledge}"\cite{polanyi-revenge} this new trend of using LLMs as knowledge sources can be viewed as a form of avenging Polanyi’s revenge (by bringing explicit knowledge back to AI systems, if only as gleaned from LLMs).\footnote{There is rich irony here: If you give what you know about a toy world to the computer, and have it solve new instances, it would be derisively called ``Good Old Fashioned AI," but if you capture all that the humanity knows about everything (as exported to the Internet), train your LLM on it,  and then ask it to provide approximate task relevant knowledge, then it becomes ``modern AI."}
Indeed, LLMs make it easy to get problem-specific knowledge, as long as we are willing to relax correctness requirements of that knowledge. In contrast to the old knowledge engineering approaches, LLMs offer this  without  making it look like we are inconveniencing any specific human (we are, instead,  just leveraging  everything humans have told each other!).  So the million dollar question for reasoning tasks becomes: “how would you do planning if you have some doddering know-it-all ready to give you any kind of knowledge?”  The LLM-Modulo framework \cite{rao-llm-modulo} is a principled method for tackling this challenge. 

To summarize, nothing that I have read, verified, or done gives me any compelling reason to believe that LLMs do reasoning/planning, as normally understood. What they do instead, armed with  web-scale training,  is a form of universal approximate retrieval, which, as I have argued, can sometimes be mistaken for reasoning capabilities. LLMs  do excel in idea generation for any task–including those involving reasoning, and as I pointed out, this can be effectively leveraged to support reasoning/planning in LLM-Modulo frameworks\cite{rao-llm-modulo}.  In other words, LLMs already  have enough amazing approximate retrieval abilities that can be gainfully leveraged, that we don't need to ascribe questionable reasoning/planning capabilities to them.\footnote{Although we focused on planning and reasoning tasks, the discussion here has implications to LLM-based code generation. There too, while LLMs can help as ``co-pilots" to human programmers, there is never any guarantee that the code they generate is correct. The fact that code generation capabilities come from {\tt github} data, that is of higher quality than the general web data, and the fact that incremental interpreters in the loop can pinpoint any syntax errors in the generated code, helps the utility of the suggested code snippets for human programmers.}

\bibliography{llmplan}

\begin{thebibliography}{10}

\bibitem{guan2023leveraging}
Lin Guan, Karthik Valmeekam, Sarath Sreedharan, and Subbarao Kambhampati.
\newblock Leveraging pre-trained large language models to construct and utilize world models for model-based task planning.
\newblock In {\em Thirty-seventh Conference on Neural Information Processing Systems}, 2023.

\bibitem{thinking-fast-slow}
Daniel Kahneman.
\newblock {\em Thinking, fast and slow}.
\newblock macmillan, 2011.

\bibitem{rao-cacm-llm}
Subbarao Kambhampati.
\newblock Language imitation games and the arrival of broad and shallow {AI}.
\newblock {\em CACM Blog}, 2021.

\bibitem{polanyi-revenge}
Subbarao Kambhampati.
\newblock Polanyi's revenge and {AI}'s new romance with tacit knowledge.
\newblock {\em Communications of the ACM}, 64(2):31--32, 2021.

\bibitem{rao-ersatz}
Subbarao Kambhampati.
\newblock {AI} as (an ersatz) natural science?
\newblock {\em Communications of the ACM}, 65(9):8--9, 2022.

\bibitem{rao-llm-modulo}
Subbarao Kambhampati, Karthik Valmeekam, Lin Guan, Kaya Stechly, Mudit Verma, Siddhant Bhambri, Lucas Saldyt, and Anil Murthy.
\newblock {LLMs} can't plan, but can help planning in {LLM-Modulo} frameworks.
\newblock {\em arXiv preprint 2402.01817}, 2024.

\bibitem{llm-tutorial}
Subbarao. Kambhampati, Karthik. Valmeekam, Matthew. Marquez, and Lin. Guan.
\newblock On the role of large language models in planning, July 2023.
\newblock Tutorial presented at the International Conference on Automated Planning and Scheduling (ICAPS), Prague.

\bibitem{nyt-travel-books}
Seth Kugel and Stephen Hiltner.
\newblock A new frontier for travel scammers: {A.I.-Generated Guidebooks}.
\newblock {\em New York Times}, August 2023.

\bibitem{stechly2023gpt}
Kaya Stechly, Matthew Marquez, and Subbarao Kambhampati.
\newblock {GPT-4 Doesn’t Know It’s Wrong: An Analysis of Iterative Prompting for Reasoning Problems}.
\newblock In {\em NeurIPS 2023 Foundation Models for Decision Making Workshop}, 2023.

\bibitem{valmeekam2023can}
Karthik Valmeekam, Matthew Marquez, and Subbarao Kambhampati.
\newblock Can large language models really improve by self-critiquing their own plans?
\newblock In {\em NeurIPS 2023 Foundation Models for Decision Making Workshop}, 2023.

\bibitem{valmeekam2023planbench}
Karthik Valmeekam, Matthew Marquez, Alberto Olmo, Sarath Sreedharan, and Subbarao Kambhampati.
\newblock Planbench: An extensible benchmark for evaluating large language models on planning and reasoning about change.
\newblock In {\em Thirty-seventh Conference on Neural Information Processing Systems Datasets and Benchmarks Track}, 2023.

\bibitem{valmeekam2023on}
Karthik Valmeekam, Matthew Marquez, Sarath Sreedharan, and Subbarao Kambhampati.
\newblock On the planning abilities of large language models - a critical investigation.
\newblock In {\em Thirty-seventh Conference on Neural Information Processing Systems}, 2023.

\bibitem{valmeekam-still-cant}
Karthik Valmeekam, Alberto Olmo, Sarath Sreedharan, and Subbarao Kambhampati.
\newblock Large language models still can't plan (a benchmark for llms on planning and reasoning about change).
\newblock {\em arXiv preprint arXiv:2206.10498}, 2022.

\end{thebibliography}
\bibliographystyle{plain}

\end{document}